\documentclass[10pt,letterpaper]{article}
\usepackage[top=0.85in,left=2.75in,footskip=0.75in]{geometry}

\usepackage{amsmath,amssymb}

\usepackage{changepage}

\usepackage[utf8x]{inputenc}

\usepackage{textcomp,marvosym}

\usepackage{cite}

\usepackage{nameref,hyperref}

\usepackage{microtype}
\DisableLigatures[f]{encoding = *, family = * }

\usepackage[table]{xcolor}

\usepackage{array}

\newcolumntype{+}{!{\vrule width 2pt}}

\newlength\savedwidth

\usepackage{setspace} 
\raggedright
\usepackage[aboveskip=1pt,labelfont=bf,labelsep=period,justification=raggedright,singlelinecheck=off]{caption}

\makeatletter
\renewcommand{\@biblabel}[1]{\quad#1.}
\makeatother

\usepackage{lastpage,fancyhdr,graphicx}
\usepackage{epstopdf}
\fancyheadoffset[L]{2.25in}
\fancyfootoffset[L]{2.25in}

\usepackage[%
sort&compress
]{cleveref}
\crefname{chapter}{section}{sections}
\Crefname{chapter}{Section}{Sections}
\Crefname{equation}{Eq}{Eq}
\Crefname{figure}{Fig}{Fig}

\usepackage{float}
\usepackage{csquotes}
\usepackage{pifont}
\usepackage{isomath}
\usepackage{bm}
\usepackage{amsmath}
\usepackage{amssymb}
\usepackage{booktabs}
\usepackage{siunitx}
\usepackage{array}
\newcolumntype{L}[1]{>{\raggedright\let\newline\\\arraybackslash\hspace{0pt}}p{#1}}
\newcolumntype{C}[1]{>{\centering\let\newline\\\arraybackslash\hspace{0pt}}p{#1}}
\newcolumntype{R}[1]{>{\raggedleft\let\newline\\\arraybackslash\hspace{0pt}}p{#1}}

\usepackage[square,sort,comma,numbers]{natbib}
\usepackage{dcolumn} 

\makeatletter
\newcolumntype{B}[3]{>{\boldmath\DC@{#1}{#2}{#3}}c<{\DC@end}}
\makeatother

\hypersetup{hidelinks,final}

\usepackage{xspace}
\newcommand\ie{i.\,e.\xspace}
\newcommand\eg{e.\,g.\xspace}
\newcommand\cf{cf.\xspace}

\newcommand{\abs}[1]{\left\lvert #1 \right\rvert}
\newcommand{\mathup}[1]{\mathrm{#1}}

\newcommand{\tablehead}{\bfseries}

\newcommand{\mcellt}[2][c]{%
	\begin{tabular}[t]{@{}#1@{}}#2\end{tabular}}

\begin{document}
\vspace*{0.2in}

\begin{flushleft}
{\Large
\textbf\newline{Statistical inferences for polarity identification\\ in natural language} 
}
\newline
\\
Nicolas Pröllochs\textsuperscript{1*},
Stefan Feuerriegel\textsuperscript{2},
Dirk Neumann\textsuperscript{1},
\\
\bigskip
\textbf{1} Chair for Information Systems Research, University of Freiburg, Freiburg, Germany
\\
\textbf{2} Chair of Management Information Systems, ETH Zurich, Zurich, Switzerland
\\
\bigskip

%
%





* Corresponding author

nicolas.proellochs@is.uni-freiburg.de (NP)

\end{flushleft}
\section*{Abstract}
Information forms the basis for all human behavior, including the ubiquitous decision-making that people constantly perform in their every day lives. It is thus the mission of researchers to understand how humans process information to reach decisions. In order to facilitate this task, this work proposes LASSO regularization as a statistical tool to extract decisive words from textual content in order to study the reception of granular expressions in natural language. This differs from the usual use of the LASSO as a predictive model and, instead, yields highly interpretable statistical inferences between the occurrences of words and an outcome variable. Accordingly, the method suggests direct implications for the social sciences: it serves as a statistical procedure for generating domain-specific dictionaries as opposed to frequently employed heuristics. In addition, researchers can now identify text segments and word choices that are statistically decisive to authors or readers and, based on this knowledge, test hypotheses from behavioral research. 




\section*{Introduction}
\label{sec:introduction}


The power of word choice and linguistic style is undisputed in the social sciences. For instance, linguistic style provides a means for deception \cite{Fuller.2013,Jensen.2013}. Likewise, marketing professionals have long understood the value of choosing the right terms when advertising products. For example, the use of technical terms facilitates the success of print advertisements \cite{Anderson.1980}. Similarly, the valence of messages helps to explain consumer behavior. Here, the use of more positive expressions in user- and marketer-generated content in social media has a clear impact on purchase decisions \cite{Goh.2013}. The subtleties of language also receive increasing attention in the financial domain. In a recent study, \cite{Bosman.2015} manipulate the tone of corporate news in a randomized controlled experiment and find that subjects expect a higher future return from a given firm when reading an article skewed towards positive language. 


Psychological research has found that negative terms, especially, are vital in forming impressions, perceptions and attributions \cite{Baumeister.2001}. For instance, subjects use more positive-emotion words in self-disclosures; yet negative-emotion words have a significantly greater impact on formed impressions \cite{Berry.1997}. Further works by Pennebaker and his colleagues shed light on the use of linguistic terms as psychological markers of personality and personal states \cite{Pennebaker.2003}. As an illustrative example, linguistic style serves as a predictor of age, gender, mood, emotion and mental health. This and other findings stem from the calculation of the occurrences of certain, aggregated word categories (\eg cognitive words, past tense, pronouns). However, though not all of the words are likely to be relevant, there is a scarcity of resources that identify the decisive entries within these categories.

While the above applications demonstrate the great importance and need for profound language understanding, the reception of individual words and their their effects on human behavior remain subject to research \cite{Agarwal.2014,Ordenes.2017,Pennebaker.2003}. When studying the reception of natural language, researchers commonly utilize a document-level score that measures the overall perception of natural language, including negative wording, tone, sentiment, moods and emotions \cite[e.\,g.][]{Bollen.2011,Pang.2008,Tetlock.2007}. However, this does not allow for a granular understanding of how individual pieces of information are perceived within narratives. In fact, understanding word choice, the perception of wording and the corresponding human responses present open questions for research, and especially in terms of rigorous statistical inferences \cite{Lim.2013}.  

Related research in the area of social sciences commonly relies on manually selected dictionaries (\eg~\cite{Shen.2015,Stieglitz.2013,Xu.2013}). Prevalent examples are the Harvard~IV psychological dictionary from the General Inquirer software or Linguistic Inquiry and Word Count~(LIWC). These contain lists of words that classify terms as either positive or negative based on human judgments, which makes them prone to severe drawbacks. Foremost, the word lists are selected \emph{ex ante} based on subjective opinions of their authors. They thus can be neither comprehensive nor as precise as statistical rigor. Furthermore among these is the fact that the labor-intensive process of their construction prevents dictionaries from being tailored to arbitrary domain-specific applications. Moreover, dictionaries rarely discriminate between different levels of positivity (or negativity), since the underlying words are merely grouped into two classes of positive and negative expressions without further weighting. 

To overcome the previous shortcomings, this paper proposes a novel approach that utilizes LASSO regularization to extract words that are statistically decisive based on an outcome variable. Examples include ordinal ratings on review portals, which summarize the connotation of user-generated comments, or the stock market reaction, which assesses investors' perceptions of financial materials. Our approach specifically builds upon these response variables as they mirror narrative content in an accurate and objective manner. Here we extend our previous work \cite{ECIS.2015} and introduce statistical inferences to identify cues that convey a positive or negative polarity. At the same time, the analysis can be replicated for the prose of arbitrary applications in order to adapt to the domain-specific particularities. 

This work immediately reveals manifold implications for social sciences and behavioral research: first, our approach offers a tailored means by which to study the perception of language and word choice through the eyes of readers and authors with statistical rigor. The results are highly interpretable and serve as an input to further hypothesis tests. 
 After all, this contributes to behavioral research by addressing the crucial question of how textual information impacts individual behavior and decision-making. 

The remainder of this paper is organized as follows. Section 2 provides background literature concerning the reception of natural language, which necessitates a statistical approach to measure the response to word choice. As a remedy, we present our LASSO-based methodology in Section 3. Subsequently, Section 4 demonstrates the value of this approach with examples from recommender systems and finance. Section 5 then provides thorough comparisons, followed by Section 6 with implications for behavioral hypothesis testing. Section 7 discusses the advantages and limitations of our method and provides detailed implications for both theory and practice.

\section*{Backgrounds}
\label{sec:related_work}

This section posits that extracting statistically relevant terms based on a decision variable is both an innovative and relevant research question to the social sciences. Therefore, we review previous works and methods concerned with measuring the reaction to word choice. We also outline how our approach differs from opinion mining, which gives a lever to measure subjective information in narrative content. 

\subsection*{Relationship to opinion mining}

Drawing inferences regarding how wording relates to a decision variable is closely related to the concept known as sentiment analysis or opinion mining. It refers to the use of natural language processing as a way to extract subjective information from narrative content. 
 The underlying methods aim at measuring the semantic orientation (\ie the positivity and negativity) of the overall text, or with respect to a particular topic or aspect \cite{Thet.2010}. The result is then either a continuous sentiment score or else a classification as positive or negative. 
 The surveys in \cite{Pang.2008} and \cite{Ravi.2015} provide a comprehensive, domain-independent overview of common methodological choices. These techniques can primarily be grouped into two categories, namely, approaches utilizing pre-defined dictionaries or machine learning. 

The former, dictionary-based approaches, mainly serve explanatory purposes, especially when a response variable is not present. They extract subjective information from the occurrences of pre-defined polarity words, which are selected ex~ante based on the intuition of experts. This creates an approach that is not only straightforward, but also produces reliable and interpretable results in various applications (\eg~\cite{Henry.2008,Tetlock.2008}). Previous research has devised several variants with different scopes and objectives (\cf next section for an overview). These dictionaries can be combined with linguistic rules that specifically account for linguistic modifiers that signal, for instance, uncertainty or activation \cite{Ordenes.2017}.

Machine learning methodologies utilize a baseline variable to train a predictive model, which is later applied to unseen documents where it should predict the semantic orientation. Previous research has tested various models, including support vector machines and artificial neural networks, that typically take (transformed) word frequencies as input (\eg~\cite{Manning.1999,Ravi.2015}). As a result, machine learning often achieves a high predictive accuracy but might suffer from overfitting. In addition, it remains a black-box with low interpretability and hardly any insights into its reasoning. 

The above approaches target applications in which whole texts are classified according to their semantic orientation. Thereby, sentiment analysis either serves explanatory or predictive purposes, which have both become prevalent in behavioral research. These methods work at document level (or aspect level); however, they cannot draw statistical inferences at word level, which is the goal of research aimed at understanding the reception of word choice at a granular level. 

\subsection*{Overview of common dictionaries}

Gaining insights into the subtle differences between word choice requires methods that analyze narrative content at a granular level. Therefore, a common strategy is to build upon manually selected dictionaries from previous research. In this vein, humans label terms as either positive and negative or, alternatively, according to other semantic, syntactic or psychological categories. \Cref{tbl:dictionaries_rw} provides an overview of prevalent dictionaries in behavioral research. For example, the Harvard~IV dictionary from the General Inquirer software comprises various psychological categories beyond positive and negative valence: \eg emotions, strength, or overstatement. LIWC was designed to identify emotion-laden writing but also measures linguistic style based on expressions that were individually assigned to over 70 linguistic dimensions by independent judges. Other dictionaries are devoted to domain-specific applications, such as the Loughran-McDonald dictionary, which consists of polarity terms found in earnings reports. With the exception of SentiStrength, SentiWordNet and QDAP, the dictionaries usually cannot differentiate between different degrees of polarity among words.

\begin{table}[H]
	\sisetup{round-mode=places,round-precision=4}
	\begin{adjustwidth}{-2.25in}{0in} 
		\centering
		\caption{\bf Common dictionaries in behavioral research.}
		{\footnotesize \singlespacing
			\begin{tabular}{lS L{3cm} l L{2cm} L{1.5cm} L{2.5cm}}
				\toprule
				\tablehead{Dictionary} & \multicolumn{1}{c}{\tablehead{Size}} & \tablehead{Categories} & \tablehead{Domain} & \tablehead{Selection Process} & \tablehead{Polarity Levels} & \tablehead{Notes} \\
				\midrule
				Diction & \num{10000} & 35 linguistic categories (\eg optimism, satisfaction, praise, blame, denial) & Politics & Expert judgment  & Binary & Accessible for purchase via the Diction software for text analysis\\
				\addlinespace
				Harvard~IV  &       \num{4206}       & 15 linguistic categories (\eg polarity, motivation, pleasure, pain, cognitive orientation) & Psychology & Expert judgment & Binary & Shipped in General Inquirer \\
				\addlinespace
				\addlinespace
				LIWC & \num{4500} & 64 linguistic dimensions (\eg polarity, part-of-speech, cognitive and psychological words) & Psychology & Independent judges & Binary &  Accessible for purchase from the LIWC text analysis software \\
				\addlinespace
				Loughran-McDonald & \num{2709} & Polarity (positive, negative) & Finance & Manual selection procedure & Binary & Based on 2of12inf dictionary\\
				\addlinespace
				QDAP & \num{6789} & Polarity (positive, negative) & General & Heuristic based on co-occurences to positive/negative seed words & Binary & Synset of WordNet \\
				\addlinespace
				SentiStrength & \num{763} & Positivity, negativity & Social media & Human judgment & Continuous rating & Derived from LIWC \\ 
				\addlinespace		
				SentiWordNet 3.0 & \num{28431} & Positivity, negativity, neutrality & General & Heuristic based on co-occurrences to positive/negative seed words & Continuous rating & Based on \num{86994} terms from WordNet \\ 
				\bottomrule
		\end{tabular}}
		\label{tbl:dictionaries_rw}
	\end{adjustwidth}
\end{table}


In order to computerize the construction of dictionaries, researchers have devised various rule-based approaches and heuristics, which are frequently refereed to as dictionary generation or lexicon creation. On the one hand, several algorithms follow a semi-supervised approach that considers word embeddings, similarity or co-occurrences between terms. For instance, SentiWordNet, as well as QDAP, starts with a small set of seed words labeled as positive or negative, based on which neighboring terms are classified \cite{Baccianella.2010}. On the other hand, some algorithms base their classifications on a response variable (the gold standard). This sounds similar to our statistical procedure, but they then propose the use of heuristics to label words depending on their appearances in documents rated with a high or low gold standard. The underlying heuristics adapt concepts from information retrieval, such as information gain, pointwise mutual information and $\chi^2$-based selection (\eg~\cite{
	Oliveira.2016}). The heuristics aim at differentiating varying degrees of sentiment strength; however, they lack statistical justification, which impairs the possibility of drawing any reliable inferences.

In addition to the above shortcomings, only a small portion of the content of the dictionaries in \Cref{tbl:dictionaries_rw} overlaps and some even contain contradictory entries. As a result, choosing the most suitable dictionary to facilitate an understanding of written information is challenging and any choice is likely to be imperfect. This is particularly relevant, since words often feature a highly domain-specific meaning. The above elaborations immediately reveal that there is no one \textquote{right} dictionary, and authors of \cite{Abbasi.2016} argue that the state-of-the-art methods for polarity scoring are subpar, which affects sentiment-related analysis and conclusions drawn from it.

\subsection*{Statistical approaches for dictionary generation}


The objective of this work is to come up with a statistical procedure that deduces the true perception of explicit and implicit polarity terms. The few existing approaches entail several statistical deficiencies. \cite{Jegadeesh.2013} count frequencies (tf-idf) of selected words and then insert them into an ordinary least squares estimation with a gold standard. However, this approach is subject to multicollinearity and, hence, the authors decided to restrict their analysis specifically to words that appear in the Loughran-McDonald finance-specific dictionary. \cite{Taddy.2013} and \cite{Taddy.2015} develop variants of multinomial regressions that can handle high-dimensional count data. However, both are limited to categorical outcome variables, which makes them infeasible in our setting. Furthermore, the multinomial regressions only work with absolute term frequencies, instead of using common weighting schemes from information retrieval (\eg tf-idf), which are often regarded as more efficient. Lastly, the underlying dimension reductions return loadings in the reduced-space, which allows for the ranking of word polarities, but lacks direct statistical interpretation (\eg standard errors). 

We later draw upon the LASSO as a procedure for extracting decisive variables. This method has been applied to textual input, but merely in predictive settings, where it serves either as tool for weighting the salience of predictive features \cite[e.\,g.][]{Joshi.2010} or black-box forecasts \cite[e.\,g.][]{Schwartz.2017}. However, we adapt it for explaining outcomes \emph{ex post}. To the best of our knowledge, it has not been combined with statistical confidence estimates or proposed as technique for measuring the reception of language. Beyond an earlier draft \cite{ECIS.2015}, the use of the LASSO has, in particular, been neither propagated as a tool for generating domain-specific dictionaries nor experimentally evaluated against manual dictionary annotations.



\subsection*{Research gap}

Altogether, we see that the above research neglects to draw rigorous statistical inferences from a comparison between word choice and the regressands. As a remedy, we develop a regularization technique to select granular polarity expressions from documents that \emph{statistically} elicit a positive or negative response. It even extracts terms that convey valence implicitly, helps to discriminate between subtle differences in polarity strength, and adapts to domain-specific particularities -- all in order to enable an in-depth analysis of the relationship between language and decisions. This ultimately contributes a better understanding of human text processing.

Previous literature has pointed out the need for understanding the reception of natural language. Related works predominantly draw on manual and labor-intensive procedures in which human judgments are assumed to reflect the ground truth. The outcome of this process usually results in a set of positive and a set of negative cues, which one refers to collectively as a dictionary. However, there is no doubt that such a setup is error-prone as perceptions of individuals and experts are eminently subjective and thus biased. These dictionaries also entail further shortcomings. First, they usually struggle to capture domain-specific characteristics. For instance, a finance-specific dictionary cannot distinguish between language describing developments in the real estate market in comparison to the linguistic style of technology firms. In addition, most dictionaries also presuppose an equal importance across all words in the same polarity group and thus do not exhaust a continuous bandwidth of sentiment levels. 



\section*{Method development}
\label{sec:methodology}

This section proposes a novel methodology by which to investigate the granular perception of natural language and to examine the textual cues that trigger decision-making. Our methodology comprises two stages, of which the first step performs several preprocessing operations to transform running text into a document-term matrix. The second step performs a variable selection to extract only the relevant terms. This essentially utilizes a LASSO regression, treating each document as an observation, while we use all words as explanatory factors explaining an exogenous response variable. We have released our method publicly in the form of an R package. The package \emph{SentimentAnalysis} is available for download via CRAN: \url{https://cran.r-project.org/package=SentimentAnalysis}.


\subsection*{Preprocessing of natural language}
\label{sec:preprocessing}

The preprocessing phase transforms the running text into a structured format that allows for further calculations. This includes a myriad of standard routines from natural language processing (\cf online appendix for details). For instance, we remove stop words without a deeper meaning and truncate inflected words to their stems~\cite{Manning.1999}. 

We then obtain frequencies $x_{d,t}$ of how often term $t$ occurs in document $d$. In order to focus only on the characteristic terms in a document, we transform the frequencies $x_{d,t}$ by using a common weighting scheme from information retrieval, namely, the term frequency-inverse document frequency or tf-idf for short \cite{Manning.1999}. Thereby, the raw frequency $x_{d,t}$ is weighted by the ratio of the total number of documents divided by the number of documents that contain the term $t$, \ie
\begin{equation}
\hat{x}_{d,t} = \text{\textit{tf-idf}}(t, d, D) = \mathit{tf}(t, d) \; \mathit{idf}(t, D) = x_{d,t} \, \log{\frac{\abs{D}}{\abs{\left\{ d \in D \mid t \in d \right\}}}},
\end{equation}
given a corpus, $D$, of documents. We have also tested different variants of using the raw term frequencies as part of our robustness checks; however, these result in a slightly lower goodness-of-fit and thus yield inferior prediction performance in both datasets.

\subsection*{Model specification}

Let $\bm{y}$ denote the gold standard that measures our response variable of interest. We now construct a linear model where the number of occurrences of individual words explains the response variable. That is, we specify a linear model 
\begin{equation}
\bm{y} = \beta_0 + \sum_{t=1}^{n}{\beta_t \bm{\hat{x}}_t} + \bm{\varepsilon}
\end{equation}
to quantify the effect of words $\bm{\hat{x}}_t = \left[ \hat{x}_{1,t}, \ldots, \hat{x}_{\abs{D},t} \right]^T$ for $t = 1, \ldots, n$ on the dependent variable $\bm{y}$ with error term $\bm{\varepsilon}$. In addition, we standardize the word variables $\bm{\hat{x}}_t$ in order to facilitate comparison between coefficients. The estimated coefficients $\beta_0, \ldots, \beta_n$ then gauge the effect of words on that gold standard. 

Estimating the above model is not trivial, since the appearance of words is likely to be highly correlated, \ie $\left| \mathup{cor}(\bm{\hat{x}}_i, \bm{\hat{x}}_j) \right| \gg 0$ for $i \neq j$.  This raises serious issues of multicollinearity and, consequently, the ordinary least squares~(OLS) estimator can be misleading. Moreover, it also results in low predictive power~\cite{Tibshirani.1996} and entails limited interpretability when facing a large number of variables~\cite{Hastie.2009}. 

In order to overcome these statistical challenges, we perform regularization via the least absolute shrinkage operator~(LASSO). Regularization can serve as a viable alternative to OLS when the number of regressors is large and highly correlated. As our main contribution to the existing body of literature on this topic, we propagate the application of this regularization approach to word frequencies in order to infer decisive words and interpret them statistically. Alternative estimators entails disadvantages, since, for instance, ridge regression and elastic net perform no variable selection and can thus not benefit from parsimonious models


\subsection*{Reasoning behind regularization}

Theory from natural language builds upon Zipf's law according to which word counts follow a power law distribution \cite{Manning.1999}. It further suggests that certain words have a potentially large impact, while a high number elicit only a marginal response \cite{Loughran.2016}. To conduct an analysis focusing on those relevant words, we need a mathematical mechanism that extracts terms deemed important. To this end, regularization is a common tool for implicit variable selection and has recently gained considerable traction in data science \cite{Varian.2014}. A common choice is the LASSO \cite{Hastie.2009,Hastie.2015,Tibshirani.1996}, since it identifies covariates that fit the data best, while simultaneously shrinking some coefficients to zero.  

The LASSO entails several properties that makes its use beneficial for estimating our model. First of all, the LASSO automatically identifies decisive variables in a linear model. It thus chooses a subset of variables and filters out non-informative model terms. In our setting, this allows us to discard words that are statistically not relevant with respect to the exogenous variable. This property of variable selection leads to parsimonious and more interpretable models. At the same time, the LASSO mitigates the issue of multicollinearity, which is present when estimating the model via ordinary least squares. Additionally, by finding a reasonable trade-off between bias and variance, it solves the problem of overfitting, which occurs if the model complexity is too high \cite{Hastie.2009,Hastie.2015,Tibshirani.1996}. 

The LASSO can be identically formalized both as an OLS estimator with an additional regularization parameter or as Bayesian model with a specific prior distribution. The LASSO has recently been extended by significance tests \cite{Lockhart.2014}. Alternatively, one can utilize standard errors from the Post-LASSO procedure \cite{Belloni.2013}.

On the whole, the LASSO specifically enables us to treat each distinct word from a corpus as a potential regressor. Its use, together with the standard errors, thereby introduces statistical inferences to natural language on a word-by-word level.

\subsection*{Statistical inferences from word choice}
\label{sec:dictionary_generation}

The LASSO incorporates an additional regularization term that penalizes non-zero coefficients~\cite{Hastie.2009,Hastie.2015,Tibshirani.1996}, given by a minimization problem 
\begin{equation}
\bm{\beta}_{\mathup{LASSO}} = \underset{\beta_0, \ldots, \beta_n}{\arg\min} \sum\limits_{i=1}^{\abs{D}} \left[ y_i - \beta_0 - \sum\limits_{t=1}^{n}{\beta_t \hat{x}_{d,t}} \right]^2 \text{ s.\,t. } \sum\limits_{t=1}^{n}{\abs{\beta_t}} \leq \lambda 
\end{equation}
with a suitable tuning parameter $\lambda$. The magnitude of the regression coefficients measures the perception of individual words statistically. 

Because of the $L_1$-penalty, the LASSO typically produces estimates in which some of the coefficients are set exactly to zero and, thereby, performs an implicit feature selection. In practice, the parameter $\lambda$ is selected using cross-validation to find a value that minimizes the error on the hold-out set. Afterwards, we re-fit the model with that specific $\lambda$ using all the observations in order to determine its coefficients. Our standard errors stem from the Post-LASSO, and allow us to make statistical tests that correspond to the use of specific words.

As a result, our procedure identifies statistically relevant words, while the corresponding coefficients measure their polarity. One major benefit of our approach is that it overcomes the problem of ex~ante selected words. Hence, we no longer run the risk of labeling words for subjective reasons or on the basis of erroneous knowledge, since all outcomes measure the influence of words on the dependent variable with statistical validation. 



\section*{Empirical results}
\label{sec:results}

This section evaluates our method with two studies from different domains: (I)~we investigate the role of word choice in recommender systems by extracting opinionated terms from user-generated reviews. (II)~We further study the impact on stock markets of the wording in financial disclosures. Subsequently, we compare the resulting word lists to the manually-selected dictionaries from previous research and show how our method can be used with higher-order word combinations to incorporate context. 

\subsection*{Study I: Opinionated terms in user-generated reviews}

\subsubsection*{Corpus with reviews.}

The first study demonstrates the added value of our approach in the domain of recommender systems, where we automatically infer terms that convey opinions. Professionals from marketing can exploit these expressions to gain insights into how people judge products or services. In related research, movie reviews represent a popular choice when it comes to studying opinion mining (\eg~\cite{Pang.2005
}). Among the reasons is that movie reviews pose a particularly difficult challenge, since they often contain a mixture of feedback, critique and summaries of movie scenes. For example, positive reviews still refer to some unpleasant scenes and negative reviews to pleasant ones.
We utilize a ubiquitous corpus of \num{5006} movie reviews from the Internet Movie Database~(IMDb), each annotated with an overall rating. The scaled dataset is available from \url{http://www.cs.cornell.edu/people/pabo/movie-review-data/}. All reviews are written by four different authors and preprocessed, \eg by removing explicit rating indicators \cite{Pang.2005}. 

\subsubsection*{Statistical inferences for polarity word scoring.}

We now extract opinionated terms from the movie reviews. The corpus contains a total number of \num{1195} word stems after preprocessing. Our methodology results in a final model with \num{549} (\SI{47.21}{\percent}) statically relevant terms. Out of these, \num{294} terms feature a positive and \num{255} a negative connotation. Unsurprisingly, the coefficients are generally small as a single word does not flip the whole meaning of the document but merely of a sentence. We report the top~15 expressions with the highest and lowest coefficients in \Cref{tbl:top_words_IMDb}. The table lists stems instead of complete words due to stemming being part of the preprocessing. We additionally calculate standard errors via the Post-LASSO~\cite{Belloni.2013}.

\Cref{tbl:top_words_IMDb} renders it possible to precisely discriminate different levels of positive and negative polarity strength. Many of the listed terms seem plausible and might be used independent of the context of a movie review, such as \emph{perfect} or \emph{bad}. These words frequently appear in sentences, such as \emph{\enquote{the story is perfect}} or \emph{\enquote{this is just a bad film}}. In addition, we observe a large number of words that are specific to the domain of motion pictures. This includes terms, such as \emph{recommend} and \emph{long}, that, for instance, occur in sentences such as \emph{\enquote{the movie was too long}}. However, other terms, such as \emph{war} (coefficient of \num{0.0041}) or \emph{crime} (coefficient of \num{0.0004}) appear unexpected at first glance. A potential reason is that these words are often related to certain actions and scenes that appeal to the audience and are -- on average -- more positively perceived than other parts in the plot. 

Furthermore, \Cref{tbl:top_words_IMDb} states, in percentage, how often each word occurs in reviews with positive or negative ratings. For instance, the term \emph{best} appears in \SI{65}{\percent} of all positive reviews and \emph{brilliant} in \SI{73}{\percent} of the cases. Yet the pure number of appearances is misleading: the term \emph{best} amounts to a much higher coefficient of \num{0.0571} compared to \num{0.0480}, thereby indicating that it expresses a more positive sentiment. We note here again that both the response variables, as well as our regressors, are standardized for easier comparisons. 

Our model features a relatively high explanatory power with an adjusted $R^2$ amounting to \num{0.5668}. We also see clear indications of multicollinearity in the model prior to performing variable selection, since 18~(\SI{1.51}{\percent}) out of all the variance inflation factors exceed the critical threshold of 4, hence, making regularization a vital ingredient of our procedure.

\begin{table}[H]
	\sisetup{round-mode=places,round-precision=4}
	\centering
\caption{\bf Empirical results of top~15 opinionated terms in movie reviews.}
	{\setlength{\tabcolsep}{3pt}\footnotesize\singlespacing\begin{tabular}{l@{\hskip .2in}S[table-format=1.4]S[table-format=1.4]S[table-format=2.2,round-precision=2]S[table-format=2.2,round-precision=2]S[table-format=2.2,round-precision=2]c}
			\toprule
			\tablehead{Word Stem} & \tablehead{Coef.} & \tablehead{{\begin{tabular}{c}Stand. Error\\(Post-LASSO)\end{tabular}}} & \tablehead{{\begin{tabular}{c}Relative\\Freq.~(\%)\end{tabular}}} & \tablehead{{\begin{tabular}{c}Positive\\Doc.~(\%)\end{tabular}}} & \tablehead{{\begin{tabular}{c}Negative\\Doc.~(\%)\end{tabular}}} & \tablehead{{\begin{tabular}{c}Harvard\\IV\end{tabular}}}\\
			\midrule
			\multicolumn{7}{c}{\textsc{Positive Terms (Top 15)}} \\
			\midrule
			great & 0.0709 & 0.0098 & 31.5821 & 66.2872 & 33.7128 & \textcircled{+} \\ 
			perfect & 0.0707 & 0.0096 & 19.1770 & 74.1667 & 25.8333 & \textcircled{+} \\ 
			excel & 0.0572 & 0.0156 & 19.6764 & 64.5685 & 35.4315 & \textcircled{+} \\ 
			best & 0.0571 & 0.0098 & 47.1634 & 64.5913 & 35.4087 & \textcircled{+} \\ 
			life & 0.0551 & 0.0102 & 49.8202 & 63.7931 & 36.2069 &  \\ 
			delight & 0.0515 & 0.0098 & 10.6872 & 76.8224 & 23.1776 & \textcircled{+} \\ 
			brilliant & 0.0480 & 0.0095 & 7.1914 & 73.0556 & 26.9444 & \textcircled{+} \\ 
			intens & 0.0469 & 0.0097 & 9.2689 & 74.3534 & 25.6466 &  \\ 
			uniqu & 0.0416 & 0.0098 & 8.3899 & 73.8095 & 26.1905 & \textcircled{+} \\ 
			recommend & 0.0393 & 0.0138 & 18.5777 & 59.7849 & 40.2151 &  \\ 
			marvel & 0.0390 & 0.0096 & 5.1938 & 79.6154 & 20.3846 & \textcircled{+} \\ 
			hilari & 0.0373 & 0.0096 & 6.9716 & 75.6447 & 24.3553 & \textcircled{+} \\ 
			easi & 0.0353 & 0.0095 & 15.4614 & 70.6718 & 29.3282 & \textcircled{+} \\ 
			matur & 0.0347 & 0.0102 & 10.4874 & 74.6667 & 25.3333 & \textcircled{+} \\ 
			fascin & 0.0346 & 0.0099 & 10.5873 & 77.7358 & 22.2642 & \textcircled{+} \\ 
			\midrule
			\multicolumn{7}{c}{\textsc{Negative Terms (Bottom 15)}} \\
			\midrule
			bad & -0.1124 & 0.0103 & 34.4986 & 47.5970 & 52.4030 & \textcircled{--} \\ 
			worst & -0.1011 & 0.0132 & 16.2605 & 52.2113 & 47.7887 & \textcircled{--} \\ 
			wast & -0.0762 & 0.0144 & 19.2769 & 52.5389 & 47.4611 & \textcircled{--} \\ 
			review & -0.0741 & 0.0169 & 53.2761 & 51.4811 & 48.5189 &  \\ 
			suppos & -0.0699 & 0.0097 & 15.6612 & 41.0714 & 58.9286 &  \\ 
			least & -0.0672 & 0.0097 & 22.6728 & 47.5771 & 52.4229 &  \\ 
			movi & -0.0671 & 0.0130 & 84.6584 & 56.0642 & 43.9358 &  \\ 
			cinematograph & -0.0538 & 0.0151 & 21.1746 & 44.8113 & 55.1887 &  \\ 
			flat & -0.0526 & 0.0096 & 6.3124 & 35.4430 & 64.5570 &  \\ 
			unfortun & -0.0512 & 0.0102 & 14.1231 & 43.4229 & 56.5771 & \textcircled{--} \\ 
			dull & -0.0483 & 0.0096 & 5.4335 & 32.7206 & 67.2794 & \textcircled{--} \\ 
			bore & -0.0483 & 0.0097 & 8.2101 & 37.2263 & 62.7737 & \textcircled{--} \\ 
			denni & -0.0468 & 0.0236 & 23.3320 & 42.2089 & 57.7911 &  \\ 
			lack & -0.0450 & 0.0097 & 16.4802 & 48.6061 & 51.3939 & \textcircled{--} \\ 
			wors & -0.0442 & 0.0097 & 7.1115 & 38.4831 & 61.5169 & \textcircled{--} \\ 
			\bottomrule
	\end{tabular}}
	\begin{flushleft}\singlespacing \emph{Notes:} This table reports the extracted terms that convey a particularly positive or negative sentiment in movie reviews. Top: the~15 most positive word stems, together with their estimated coefficient. Standard errors are calculated via the Post-LASSO~\cite{Belloni.2013}. Bottom: the~15 most negative word stems. In addition, we provide the relative frequency within the corpus, as well as the ratio of positive and negative documents that contain each word. The last column show the overlap with the Harvard~IV psychological dictionary. The symbol \textquote{\textcircled{+}} indicates terms that appear in the positive word list and \textquote{\textcircled{--}} in the negative word list of this dictionary. The complete list with all 549 stems is given in the supplementary materials.
\end{flushleft}
\label{tbl:top_words_IMDb}
\end{table}

\Cref{tbl:top_words_IMDb} also compares the inferred polarity score to expert judgments. Evidently, there is a considerable number of opinionated terms that are not covered by dictionary word lists. Among the 15 most positive words, for example, only 12 have found their way in the Harvard~IV psychological dictionary, whereas this is true for only 8 of the 15 most negative terms. We later detail the overlap for the complete list of terms in Section 5, finding only a minor consensus of \SI{40.44}{\percent}. This stems from the fact that authors commonly utilize implicit polarity words to express their opinions, which are not included in psychological dictionaries. This highlights the shortcomings of human dictionaries and provides strong evidence that authors convey their message by utilizing different and highly domain-specific wording to communicate their opinion.

\subsection*{Study II: Impact of wording on financial markets}

\subsubsection*{Financial corpus.}

Our second study demonstrates the reception of language in regulatory Form~8-K filings from the United States. These inform investors about important corporate events, such as management changes, the departure of directors, bankruptcy, layoffs, and other events deemed significant. Form~8-K filings are quality-checked by the Securities and Exchange Commission~(SEC) to ensure that the content meets formal requirements. These reports are of high relevance for the stock market and communicate very informative material \cite{Carter.1999}; this suggests a strong relationship between their content and market responses. 

Our filings (including amendments) span the years 2004 to 2013, originating from the EDGAR website of the SEC. The complete sample consists of \num{901133} filings, which then undergo several filtering steps. First, we select only filings from firms whose stocks were publicly traded on the New York Stock Exchange~(NYSE). Second, in order to gain information about the stock market reaction, we remove filings for which we are not able match the SEC~CIK numbers to Thomson Reuters Datastream (from which all financial data is retrieved). Consistent with prior research, we exclude filings that contain fewer that \num{200} words and penny stocks below~\$5 per share \cite{Loughran.2011}. These filtering steps then result in a final corpus of \num{76717} filings.

We measure the stock market reaction subsequent to a disclosure by the abnormal return of the corresponding company, since it corrects the nominal return for concurrent market movements. 
In short, we implement a market model that assumes a stable linear relation between market return and normal return. We model the market return using a stock market index, namely, the NYSE Composite Index, along with an event window of 10~trading days prior to the disclosure. The supplementary materials provide a thorough explanation of this approach. 

\subsubsection*{Statistical inferences for word reception.}

We now report the cues that are relevant for the decision-making of investors when reading financial materials. Our approach selects a total of \num{172} statistically relevant terms, out of \num{1724} entries in the preprocessed corpus, \ie \SI{9.98}{\percent}. Out of this subset, \num{82} entries are linked to a positive stock market response, \num{90} word stems to a decreasing firm valuation. Such a relatively small subset of decisive terms is in line with the suggestion from Zipf's law \cite{Loughran.2016,Manning.1999}. We observe generally smaller coefficients as compared to our first study with movie reviews. This is not an unexpected result, since the average length of financial filings (\num{3473} words) is higher than that of reviews (\num{1066} words). Hence, the proportional influence of a single word as measured by the magnitude of its coefficient is smaller. \Cref{tbl:top_words_8K} reports the 15 words with the highest and lowest coefficients based on our procedure, for which we again provide only stemmed words due to our preprocessing. As before, we additionally calculate standard errors via the Post-LASSO. The complete list is provided in the supplements.

Similarly to the previous corpus, we observe several terms that are specific to the given domain of financial reporting, \eg \emph{improv}, \emph{strong} \emph{payrol} and \emph{lower}. These words crop up, for instance, in sentences such as \emph{\enquote{the strong business development was sustainably confirmed}}. In contrast, we also find unexpected outcomes, which appear predominantly in the negative list. Examples include \emph{although} (standardized coefficient of \num{-0.0036}) and \emph{however} (standardized coefficient of \num{-0.0015}). Most likely, these cues convey uncertainty, attenuate other statements or overturn earlier expectations. 

Overall, the current model features a lower explanatory power 
when compared to the previous model based on user-generated reviews. We expected such an outcome, since previous work has found that very few variables can predict stock returns in efficient markets \cite{Tetlock.2008}. In addition, we see again strong evidence of multicollinearity, since 24~(\SI{1.39}{\percent}) of the variance inflation factors in the full model before variable selection exceed the critical threshold of~4. This stresses once more the need for regularization in our approach.

\begin{table}[H]
		\begin{adjustwidth}{-2.25in}{0in} 
	\sisetup{round-mode=places,round-precision=4}
	\centering
\caption{\bf Empirical results of top~15 polarity expressions in financial filings.}
	{\setlength{\tabcolsep}{3pt}\footnotesize\singlespacing\begin{tabular}{l@{\hskip.2in}S[table-format=1.4]S[table-format=1.4]S[table-format=2.2,round-precision=2]S[table-format=2.2,round-precision=2]S[table-format=2.2,round-precision=2]cc}
			\toprule
			\tablehead{Word Stem} & \tablehead{Coef.} & \tablehead{{\begin{tabular}{c}Stand. Error\\(Post-LASSO)\end{tabular}}} & \tablehead{{\begin{tabular}{c}Relative\\Freq.~(\%)\end{tabular}}} & \tablehead{{\begin{tabular}{c}Positive\\Doc.~(\%)\end{tabular}}} & \tablehead{{\begin{tabular}{c}Negative\\Doc.~(\%)\end{tabular}}} & \tablehead{{\begin{tabular}{c}Harvard\\IV\end{tabular}}}  & \tablehead{{\begin{tabular}{c}Loughran-\\McDonald\end{tabular}}} \\
			\midrule
			\multicolumn{8}{c}{\textsc{Positive Terms (Top 15)}} \\
			\midrule
			improv & 0.0325 & 0.0045 & 37.2692 & 49.7026 & 50.2974 & \textcircled{+} & \textcircled{+} \\ 
			rais & 0.0160 & 0.0038 & 11.0629 & 51.1668 & 48.8332 & \textcircled{--} &  \\ 
			strong & 0.0144 & 0.0045 & 28.4175 & 50.3210 & 49.6790 &  & \textcircled{+} \\ 
			increas & 0.0113 & 0.0051 & 60.5142 & 49.1647 & 50.8353 &  &  \\ 
			facil & 0.0106 & 0.0039 & 35.2824 & 49.2913 & 50.7087 &  &  \\ 
			waiver & 0.0095 & 0.0039 & 15.3299 & 48.3336 & 51.6664 &  &  \\ 
			stronger & 0.0080 & 0.0039 & 5.4402 & 50.5839 & 49.4161 &  & \textcircled{+} \\ 
			vacat & 0.0076 & 0.0037 & 5.5754 & 49.7212 & 50.2788 &  &  \\ 
			repurchas & 0.0074 & 0.0039 & 22.9706 & 50.0265 & 49.9735 &  &  \\ 
			favor & 0.0073 & 0.0040 & 25.4467 & 49.7105 & 50.2895 & \textcircled{+} & \textcircled{+} \\ 
			consumm & 0.0067 & 0.0040 & 15.1299 & 48.4277 & 51.5723 & \textcircled{+} &  \\ 
			annum & 0.0056 & 0.0039 & 9.2707 & 47.9953 & 52.0047 &  &  \\ 
			avoid & 0.0051 & 0.0037 & 11.5549 & 48.8244 & 51.1756 & \textcircled{--} &  \\ 
			payrol & 0.0049 & 0.0037 & 6.6918 & 49.3234 & 50.6766 &  &  \\ 
			middl & 0.0046 & 0.0037 & 5.1591 & 49.1485 & 50.8515 &  &  \\ 
			\midrule
			\multicolumn{8}{c}{\textsc{Negative Terms (Bottom 15)}} \\
			\midrule
			declin & -0.0204 & 0.0045 & 23.5855 & 48.6476 & 51.3524 & \textcircled{--} & \textcircled{--} \\ 
			negat & -0.0162 & 0.0040 & 20.0295 & 47.8507 & 52.1493 & \textcircled{--} & \textcircled{--} \\ 
			lower & -0.0138 & 0.0047 & 27.2673 & 48.7211 & 51.2789 & \textcircled{--} &  \\ 
			experienc & -0.0117 & 0.0038 & 12.1699 & 47.9343 & 52.0657 &  &  \\ 
			delay & -0.0091 & 0.0038 & 18.7157 & 47.6493 & 52.3507 & \textcircled{--} & \textcircled{--} \\ 
			broad & -0.0063 & 0.0038 & 11.4562 & 48.2185 & 51.7815 &  &  \\ 
			advertis & -0.0056 & 0.0042 & 8.8557 & 48.3516 & 51.6484 &  &  \\ 
			project & -0.0055 & 0.0041 & 36.8759 & 48.8436 & 51.1564 &  &  \\ 
			pressur & -0.0055 & 0.0038 & 9.4221 & 48.9600 & 51.0400 &  &  \\ 
			now & -0.0054 & 0.0040 & 27.1362 & 48.8021 & 51.1979 &  &  \\ 
			challeng & -0.0054 & 0.0039 & 15.5800 & 48.4081 & 51.5919 & \textcircled{--} & \textcircled{--} \\ 
			offer & -0.0052 & 0.0045 & 40.3265 & 48.7565 & 51.2435 & \textcircled{+} &  \\ 
			depreci & -0.0051 & 0.0052 & 23.1111 & 48.5876 & 51.4124 & \textcircled{--} &  \\ 
			impact & -0.0041 & 0.0046 & 39.7913 & 48.6175 & 51.3825 &  &  \\ 
			weak & -0.0039 & 0.0038 & 8.5841 & 48.3546 & 51.6454 & \textcircled{--} & \textcircled{--} \\ 
			\bottomrule
	\end{tabular}}
\begin{flushleft}\singlespacing \emph{Notes:} This table reports verbal expressions that convey positive and negative information in financial disclosures (Form~8-K filings). Top: 15~most positive word stems, together with their estimated coefficient. Standard errors are calculated via the Post-LASSO~\cite{Belloni.2013}. Bottom: the~15 most negative word stems. In addition, we provide the relative frequency in financial filings, as well as the ratio of documents with a positive or negative market response. The last columns shows the agreement between our statistical inferences and two common dictionaries based on human annotations, namely, the Harvard~IV psychological and Loughran-McDonald finance-specific dictionary. The symbol \textquote{\textcircled{+}} indicates terms that appear in the respective positive word list, \textquote{\textcircled{--}} in the negative one. The complete table with all \num{172} entries is given in the supplements.
\end{flushleft}
\label{tbl:top_words_8K}
\end{adjustwidth}
\end{table}

In addition, \Cref{tbl:top_words_8K} compares the inferred polarity scores to the classifications from psychological and finance-specific dictionaries. As suggested by \cite{Loughran.2016}, we observe that dictionary labels deviate extensively from the true perception of stock market investors. From the 15 most positive words, only 5 words are also contained in the Harvard~IV psychological dictionary, whereas this is true for 8 of the negative terms. Similarly, the Loughran-McDonald dictionary contains only 4 out of the 15 most positive words and 5 out of the 15 most negative words. A detailed comparison with other dictionaries is provided as part of our robustness checks in Section 5. Overall, this indicates that the human experts do not accurately judge the true reception of natural language in the financial domain.

\subsection*{Comparison to dictionaries from human selection}

We now compare the results of our statistical inferences to the manually selected dictionaries from previous research. For this purpose, \Cref{tbl:dictionary_overlap} details the number of overlapping terms and compares to what extent classifications agree. In addition, we present the inter-rater reliability (\ie the concordance with our statistical inferences) in terms of Krippendorff's alpha coefficient \cite{Krippendorff.2013}. Here, a reliability value of 1 indicates a perfect overlap between the classifications in positive and negative groups, whereas a value of 0 denotes that human dictionaries and our statistical inferences are statistically unrelated.

\begin{table}[H]
	\begin{adjustwidth}{-2.25in}{0in} 
		\sisetup{round-mode=places,round-precision=4}
		\centering
		\caption{\bf Comparison of human classifications to statistical inferences.}
		{\setlength{\tabcolsep}{3pt}\footnotesize\singlespacing\begin{tabular}{l@{\hskip .2in} S[table-format=5.0]S[table-format=3.0]S[table-format=1.4] @{\extracolsep{0.3cm}} S[table-format=3.0]S[table-format=1.4] @{\extracolsep{0.3cm}}  D{.}{.}{5.7} S[table-format=1.4]}
				\toprule
				& \multicolumn{1}{c}{\tablehead{Size}} & \multicolumn{2}{c}{\tablehead{Overlapping Terms}} & \multicolumn{2}{c}{\tablehead{Consensus Classification}} & \multicolumn{1}{c}{\tablehead{Correlation}} & \tablehead{Reliability} \\
				\cmidrule{3-4}\cmidrule{5-6} && {Count} & {Share}                                 & {Count} & {Share}                                        &  \\ \midrule
				\multicolumn{8}{c}{\textsc{Study I: Movie Reviews}}                                                                                 \\ \midrule
				Harvard~IV                   & 4206 & 222     & 0.4044                                  & 138     & 0.6216                                         & 0.3236^{***}                                & 0.2246                  \\
				Henry                        & 190 & 26      & 0.0474                                  & 20      & 0.7692                                         & 0.5593^{**}                                 & 0.5446                  \\
				Loughran-McDonald            & 2709 & 73      & 0.1330                                  & 45      & 0.6164                                         & 0.4303^{***}                                & 0.2311                  \\
				SentiWordNet                 & 28431 & 440     & 0.8015                                  & 246     & 0.5591                                         & 0.2649^{***}                                & 0.1001                  \\
				QDAP                         & 6789 & 176     & 0.3206                                  & 114     & 0.6477                                         & 0.3638^{***}                                & 0.2863                  \\ \midrule
				\multicolumn{8}{c}{\textsc{Study II: Financial Filings}}                                                                               \\ \midrule
				Harvard~IV                   & 4206 & 55      & 0.3198                                  & 34      & 0.6182                                         & 0.2742^{*}                                  & 0.2270                  \\
				Henry                        & 190 & 21      & 0.1221                                  & 19      & 0.9048                                         & 0.6333^{**}                                 & 0.8102                  \\
				Loughran-McDonald            & 2709 & 20      & 0.1163                                  & 18      & 0.9000                                         & 0.6433^{**}                                 & 0.8030                  \\
				SentiWordNet                 & 28431 & 118     & 0.6860                                  & 69      & 0.5847                                         & 0.2089^{*}                                  & 0.1715                  \\
				QDAP                         & 6789 & 40      & 0.2326                                  & 28      & 0.7000                                         & 0.4524^{**}                                 & 0.3939                  \\ \midrule
				\multicolumn{8}{r}{\footnotesize{Signif.: $^{***} 0.001$, $^{**} 0.01$, $^* 0.05$.}} \\ \bottomrule
		\end{tabular}}
		\label{tbl:dictionary_overlap}
		\begin{flushleft}\singlespacing \emph{Notes:} This table compares common, human-generated word lists to extracted terms based on our statistical inferences. We omitted LIWC and Diction, since these are commercial products with proprietary dictionaries. When computing correlation coefficients and reliability scores, we exclude non-overlapping terms and count binary dictionary entries with a negative label as \num{-1} and positive ones as \num{1}. Reliability (\ie the concordance with our statistical inferences) is measured in terms of Krippendorff's alpha coefficient \cite{Krippendorff.2013}.
		\end{flushleft}
	\end{adjustwidth}
\end{table}

The results demonstrate that the ex ante selected dictionaries show only a small overlap with the word lists from our statistical procedure. In the case of movie reviews, only \num{222} out of \num{549}~(\ie \SI{40.44}{\percent}) extracted words have found their way into the Harvard~IV dictionary that is frequently utilized in IS and behavioral research. Out of these, only \SI{62.16}{\percent} actually exhibit the same polarity direction. This is in line with our in-depth investigations, since many negative expressions from this dictionary feature a positive connotation in the context of movie evaluations. Psychological dictionaries classify words, such as such as \emph{crime}, \emph{force} or \emph{war}, in the negative list, while, in film reviews, these often refer in a positive sense to the suspense in certain scenes. Unsurprisingly, we find the highest number of overlapping terms in the dictionary that includes the most entries, \ie the SentiWordNet. However, this dictionary shows the lowest reliability~(\num{0.10}) and correlation~(\num{0.26}) with our statistical inferences. In contrast, the highest reliability~(\num{0.54}) and correlation~(\num{0.56}) is achieved by the Henry dictionary which, however, consists of a mere \num{190} entries, resulting in a minor overlap of \num{26} words. 

We observe similar results for our financial disclosures, where \num{55} out of \num{172} extracted words~(\ie \SI{31.98}{\percent}) also appear in the Harvard~IV dictionary. Out of these, \SI{61.82}{\percent} feature the same direction. Overall, we find a correlation of \num{0.27} between the estimated coefficients and the binary Harvard~IV dictionary (encoded as $\pm 1$). Even the dictionaries that were specifically designed for financial reports reveal large deviations from the statistical inferences. We observe only a total number of \num{21} overlapping terms for the Henry dictionary, and \num{20} for the Loughran-McDonald dictionary. Nonetheless, compared to psychological dictionaries, we see that the finance-specific dictionaries are indeed more accurate in measuring the reception of words in financial disclosures. For example, the Loughran-McDonald dictionary shows a consensus classification of \SI{90.48}{\percent} and a correlation of \num{0.64} with our statistical inferences. Moreover, finance-specific dictionaries also yield the highest  reliability. For example, the Henry dictionary shows a Krippendorff's alpha coefficient of \num{0.8102} (compared to \eg \num{0.2270} for the Harvard~IV).

\Cref{tbl:dictionary_overlap} identifies a consistent disagreement between human classification and statistical selection. Although most ex ante dictionaries feature a large volume of words, many statistically relevant terms are not included. In addition, overlapping terms show a relatively low correlation that is, in some cases, only significant at the \SI{5}{\percent} level. As a consequence, misclassification and the erroneous exclusion of words limit the suitability of ex ante dictionaries. 

The aforementioned dictionaries have frequently been utilized also in predictive settings and we thus also compare the out-of-sample performance of the above dictionaries with our method. We briefly outline the results here, while we provide further statistics and elaboration in our supplementary materials. In short, our method outperforms all of the investigated dictionaries for both, movie reviews and financial disclosures. In the case of movie reviews, the best performing dictionary (Harvard~IV) results in a \SI{90.66}{\percent} higher mean squared error compared to the LASSO. We observe a similar pattern for financial disclosures. 
These results thus reinforce our previous finding that manually selected dictionaries deviate from true perception.

\subsection*{Statistical inferences with word phrases}

Human-generated dictionaries commonly categorize only isolated words without incorporating any contextual information. However, the position of a word in a sentence is likely to contribute to the meaning and the overall interpretation. Consequently, related research attempts to work with higher-order word combinations, \ie so-called $n$-grams. However, findings indicate mixed results regarding the extent to which their inclusion improves performance. 
Expert dictionaries refrain from labeling word pairs, since it requires considerable manual labor. Similarly, heuristics for dictionary creation are also rarely designed to process $n$-grams. This is in contrast to our statistical procedure, which works effortlessly with $n$-grams as the corresponding frequencies are simply inserted in the variable selection procedure. These benefits become particularly evident when considering the sheer number of input variables (\num{2971} bigrams for financial filings and \num{1059} bigrams for movie reviews). Such large numbers of highly correlated predictors would imply serious overfitting issues for almost any type of statistical model without variable selection.

\Cref{tbl:results_ngrams} compares the results from using $n$-grams. First of all, we observe fewer relevant bigrams than unigrams. In the case of unigrams, our method extracts \num{549} relevant terms from the movie reviews and \num{172} from the financial corpus, while using bigrams results in a total number of \num{442} terms for movie reviews and \num{51} for financial filings. We provide the complete lists of extracted phrases in the supplementary materials due to space limitations, but summarize a few intriguing insights here. For instance, the bigram with the highest positive coefficient in the review corpus is \emph{best film}, while the most negative bigrams are \emph{bad movie} and \emph{waste time}. 

According to \Cref{tbl:results_ngrams}, we also observe a drop in the adjusted $R^2$ for both corpora. In the case of movie reviews, the adjusted $R^2$ declines from \num{0.5668} for unigrams to \num{0.3184} for bigrams due to its penalty on the degrees-of-freedom. We observe a similar pattern for our financial corpus. Here, the adjusted $R^2$ decreases from \num{0.0079} for unigrams to \num{0.0036} for bigrams. Finally, we also tested a configuration that incorporates both unigrams and bigrams. While this approach yields the highest fit for the review corpus, we observe a slightly inferior goodness-of-fit for the financial corpus. Altogether, this shows that our method is not limited to single terms, but also serves as an appropriate tool to study the influence of higher-order word combinations, and even phrases, on a response variable.

\begin{table}[H]
	\sisetup{round-mode=places,round-precision=4}
	\sisetup{free-standing-units}
	\centering
\caption{\bf Summary statistics of statistical inferences with word tuples.}
	{\setlength{\tabcolsep}{3pt}\footnotesize\singlespacing\begin{tabular}{l@{\hskip .2in}S[table-format=4.4,table-align-text-post=false]S[table-format=4.4,table-align-text-post=false]}
			\toprule
			&\multicolumn{1}{c}{\textsc{\mcellt{Study I:\\ Movie Reviews}}} & \multicolumn{1}{c}{\textsc{\mcellt{Study II:\\ Financial Filings}}}\\
			\midrule
			\multicolumn{3}{l}{\underline{Bigrams}} \\
			Regressors before regularization & 1059 & 2971\\
			Extracted terms & 442.0 & 47.0 \\
			Ratio of extracted terms & 41.74\,\% & 1.58\,\% \\
			Positive terms & 234 & 19\\
			Negative terms & 208 & 28\\
			Ratio positive terms & 52.94\,\% & 40.43\,\%\\
			Ratio negative terms & 47.06\,\% & 59.58\,\%\\
			\addlinespace
			Adjusted $R^2$ & 0.3184 & 0.0036\\
			Correlation between model estimate &  0.63 & 0.08\\
			and gold standard \\
			\addlinespace
			\multicolumn{3}{l}{\underline{Bigrams \& unigrams}} \\
			Regressors before regularization & 2254 & 4695\\
			Extracted terms & 798 & 132 \\
			Ratio of extracted terms & 35.40\,\% & 2.811\,\% \\
			Positive terms & 394 & 62\\
			Negative terms & 404 & 70\\
			Ratio positive terms & 49.37\,\% & 46.97\,\%\\
			Ratio negative terms & 50.63\,\% & 53.03\,\%\\
			\addlinespace
			Adjusted $R^2$ & 0.6126 & 0.0072\\
			Correlation between model estimate & 0.83 & 0.11\\
			and gold standard \\
			\bottomrule
	\end{tabular}}
	\begin{flushleft}\singlespacing \emph{Notes:} The table compares our statistical inferences for different inputs, consisting of bigrams and the combination of unigrams and bigrams. These are evaluated in terms of goodness-of-fit and by comparing the number of selected entries. The complete lists of extracted variables and their coefficients are given in the supplements.
\end{flushleft}
\label{tbl:results_ngrams}
\end{table}

\section*{Implications for hypothesis testing using natural language}
\label{sec:hypothesis_testing}

Our method presents also a valuable tool for analyzing behavioral research questions. This section demonstrates two applications that allow for the testing of hypotheses with focus on word choice. 

\subsection*{Placement of negative information in movie reviews}

We utilize our method to test where authors place negative statements in their reviews. Writers might start with negative thoughts, as suggested by the law of primacy in persuasion. On the other hand, one might be inclined to instead utilize the regency effect, according to which arguments presented last garner more attention. Given the overall movie rating, we can evaluate where authors place negative information when composing movie reviews, \ie do they generally introduce negative aspects at the beginning or rather at the end? 

\noindent
\textsc{Hypothesis:} \emph{Negative information is more likely to be placed at the end than at the beginning of a review.}

In order to test this hypothesis, we compute the sentiment of the first and second half of each review by summing over products of coefficient and weighted term frequency. We refer to them as $\mu_1$ and $\mu_2$ respectively. Summary statistics of $\mu_1$, $\mu_2$ and the document sentiment $\mu$ are shown in the first panel of \Cref{tbl:results_hypothesis_imdb}. In addition, we present the same statistics for reviews that are filtered for a positive (Panel~II) or negative (Panel~III) gold standard only. We then test the null hypotheses 
\begin{align*}
H_0: & \quad \mu_1 < \mu_2 && \text{(law of primacy applied to negative content)} , \text{ and} \\
H_0: & \quad \mu_1 > \mu_2 && \text{(regency effect for negative content)} 
\end{align*}
respectively. 

According to our results, the second half of movie reviews generally conveys a more negative tone than the first half. The mean sentiment in the first half amounts to $\mu_1 = \num{0.1025}$, whereas it is $\mu_2 = \num{0.0578}$ for the remainder part. The corresponding difference $\mu_1 - \mu_2$ between both sentiment values is statistically significant at the \SI{0.1}{\percent} significance level when performing a two-sided Welch $t$-test (test statistic of \num{15.06}). The results in Panel~II and III follow a similar picture. For instance, Panel~III shows that the first half of negatively rated movie reviews yields a negative sentiment of $\mu_1 = \num{-0.0098}$ on average, while the second half results in an even more negative sentiment of $\mu_2 = \num{-0.0930}$. This difference is also significant at the \SI{0.1}{\percent} significance level with a $t$-value of \num{19.26}. In Panel~II, we observe a similar pattern for reviews with positive ratings ($t$-value of \num{6.70}). We thus accept our hypothesis regarding the presence of a regency effect. This result also coincides with psychological research according to which senders of information are more likely to place negative content at the end \cite{Legg.2014}, but, in contrast, our evidence is collected outside of an artificial laboratory setting, as it stemms from actual human communication.

\begin{table}[H]
		\begin{adjustwidth}{-2.25in}{0in} 
	\sisetup{round-mode=places,round-precision=4}
	\centering
	\caption{\bf Summary statistics for hypothesis testing with movie reviews.\label{tbl:results_hypothesis_imdb}}
	{\setlength{\tabcolsep}{3pt}\footnotesize\singlespacing\begin{tabular}{l@{\hskip .2in}SSS@{\hskip .1in} @{\hskip .1in}SSS@{\hskip .1in} @{\hskip .1in}SSS}
			\toprule
			&\multicolumn{3}{c}{\textsc{\mcellt{Panel I: \\All Reviews}}} & \multicolumn{3}{c}{\textsc{\mcellt{Panel II: \\Positive Rating}}} & \multicolumn{3}{c}{\textsc{\mcellt{Panel III: \\Negative Rating}}} \\
			\cmidrule(l){2-4}\cmidrule(l){5-7}\cmidrule(l){8-10}
			&\multicolumn{1}{c}{$\boldsymbol{\mu_1}$} & \multicolumn{1}{c}{$\boldsymbol{\mu_2}$} & \multicolumn{1}{c}{$\boldsymbol{\mu}$}&\multicolumn{1}{c}{$\boldsymbol{\mu_1}$} & \multicolumn{1}{c}{$\boldsymbol{\mu_2}$} & \multicolumn{1}{c}{$\boldsymbol{\mu}$}&\multicolumn{1}{c}{$\boldsymbol{\mu_1}$} & \multicolumn{1}{c}{$\boldsymbol{\mu_2}$} & \multicolumn{1}{c}{$\boldsymbol{\mu}$}\\
			\midrule
			Mean & 0.1025 & 0.0578 & 0.1604 & 0.1626 & 0.1402 & 0.3027 & -0.0098 & -0.0930 & -0.1028 \\ 
			\midrule
			Min. & -0.4159 & -0.5922 & -0.8903 & -0.2020 & -0.3977 & -0.2947 & -0.4159 & -0.5922 & -0.8903 \\ 
			{25\,\% Quantile} & 0.0159 & -0.0517 & -0.0173  & 0.0825 & 0.0507 & 0.1539 & -0.0729 & -0.1639 & -0.2028 \\ 
			Median & 0.0996 & 0.0528 & 0.1476 & 0.1557 & 0.1342 & 0.2896 & -0.0021 & -0.0849 & -0.0947 \\ 
			{75\,\% Quantile} & 0.1853 & 0.1646 & 0.3331  & 0.2325 & 0.2251 & 0.4302  & 0.0587 & -0.0178 & -0.0014 \\ 
			Max. & 0.7336 & 0.7655 & 1.3848 & 0.7336 & 0.7655 & 1.3848  & 0.3624 & 0.3322 & 0.3687 \\ 
			Std. Dev. & 0.1340 & 0.1615 & 0.2569 & 0.1175 & 0.1357 & 0.2066  & 0.1060 & 0.1123 & 0.1604 \\ 
			Skewness & 0.1717 & 0.1535 & 0.2202 & 0.4439 & 0.3028 & 0.5058 & -0.2684 & -0.3898 & -0.3062 \\ 
			Kurtosis & 0.6052 & 0.2682 & 0.2204 & 0.6407 & 0.5611 & 0.6167 & 0.5321 & 0.7152 & 0.6622 \\ 
			\bottomrule
	\end{tabular}}
	\begin{flushleft}\singlespacing \emph{Notes:} Panel~I compares the sentiment of the first ($\mu_1$) and second half ($\mu_2$) of movie reviews, as well as the overall sentiment $\mu$. The additional panels present the same statistics for reviews with positive (Panel~II) or negative (Panel~III) gold standard only.
\end{flushleft}
\end{adjustwidth}
\end{table}

\subsection*{Response of financial markets to non-informative wording}

In our second application of hypothesis testing, we examine to what extent financial markets trade upon non-informative wording. Previous works have established a robust market response to fact-related information encoded in written materials, which is primarily measured by using the positive and negative word lists from Loughran-McDonald or Harvard~IV. Yet it is unclear how the remaining words -- which are not deemed as either positive or negative from a external standpoint and which we refer to as non-informative -- are processed by markets. Consistent with classical economic theory, we expect that investors ignore these terms and, instead, solely focus on essential, fact-related information, \ie clearly positive and negative cues. 

\noindent
\textsc{Hypothesis:} \emph{Financial markets are not distracted by the wording in corporate communication that falls outside the clearly delineated categories of positive and negative.} 

Interestingly, we present empirical results in the following section which reject the above hypothesis and suggest the opposite. The extracted words from \Cref{tbl:top_words_8K} list the polarity terms that are statistically relevant for the investment decisions of traders. However, most of them are not necessarily classified as positive or negative according to the Harvard IV psychological or Loughran-McDonald finance-specific dictionary. We thus test our hypothesis by grouping all words into two categories according to the previous dictionaries: one group contains all words that are labeled as either positive or negative. This group represents all terms that feature an explicit, fact-based statement. The remaining entries form a group that can be characterized as non-informative wording. For instance, the latter contains entries, such as \emph{although} and \emph{however}. We find that the perception of investors depends on many terms that feature no explicit positive or negative statement polarity. According to the Harvard~IV dictionary, only \SI{31.97}{\percent} of the extracted words can be associated with a fact-based meaning, whereas \SI{68.03}{\percent} of the extracted words are expected not to contribute to the informative content. The Loughran-McDonald dictionary presents a similar picture. Here, the fact-based group contains \SI{11.63}{\percent} of all extracted words, while the remaining \SI{88.37}{\percent} can be regarded as non-informative wording. 

Finally, we perform an $F$-test to validate whether the subset of words that are neither labeled as positive nor negative has a combined effect on stock returns. In the case of the Harvard~IV dictionary, this results in an $F$-statistic of \num{5.37}, which is statistically significant at the \SI{0.1}{\percent} level. Similarly, the $F$-statistic for the Loughran-McDonald dictionary numbers to \num{5.48}, which is also significant the \SI{0.1}{\percent} level. We must thus reject our hypothesis and provide evidence that expressions deemed as non-informative wording by previous research have a statistically significant effect on financial markets.

\section*{Discussion}
\label{sec:discussion}

In the following, we discuss the implications of our research method as it not only improves understanding of natural language but also enables intriguing inferences in behavioral sciences. Furthermore, our research is highly relevant for practitioners seeking to operationalize natural language in Information Systems.

\subsection*{Implications for behavioral sciences}



Understanding decision-making and providing decision support both increasingly rely upon computerized natural language processing. In contrast to many black-box methods from the domain of machine learning, our methodology provides a vehicle for content analysis and opinion mining that is fully comprehensible for deep insights. Specifically, it allows one to maintain high interpretability as it explains an effect in terms of the presence of individual words. It thus allows researchers to dissect the relationship between natural language and a given outcome variable. In addition, our approach goes beyond pre-defined dictionaries that classify words into groups of positive and negative words as we assign individual word weights to each word, thereby accounting for differences in the valence levels of words of the same polarity class.

Our results indicate that common, manually selected dictionaries from the literature, such as the Harvard~IV psychological dictionary, are neither complete nor adequate for arbitrary domains. For instance, in the area of finance, they classify words as positive that are not necessarily interpreted positively by investors. To overcome these previous limitations, our methodology provides a means by which to automate the process of dictionary generation. 
 Altogether, our study thus provides evidence that applications of dictionary-based sentiment analysis can be significantly improved when adapting the dictionaries to the corresponding domain.




\subsection*{Applications}

Analyzing the perceptions of word choice and understanding the response to natural language on a granular level can yield new insights in a large number of use cases. In the following points, we illustrate prominent applications in the areas of both practice and research:
\begin{quote}\begin{itemize}
		\item \textbf{Recommender systems.} Recommender systems support users by predicting their rating or preference towards products or services, and, similarly, product reviews on web platforms guide individuals considering purchases. Yet it is unclear which expressions actually convey an opinion, even though this would allow for a better understanding of how judgments are formed. 
 Our statistical inferences aid enterprises in identifying success factors of products, while they present researchers new opportunities to study behavioral theories at word level. 
		\item \textbf{Social and behavioral sciences.} In the context of social interactions, it is highly relevant to understand how humans express and perceive information in natural language. Our methodology helps to answer various questions, such as which wording drives word-of-mouth. Moreover, it enables the identification of word choices that convey information regarding personality and psychology or linguistic cues that are linked to deception in human communication. 
		\item \textbf{Finance.} Before exercising ownership in stocks, investors often consult financial disclosures and pay especially close attention to their soft content, such as linguistic style.
	To analyze the language in financial materials, researchers, investors and automated traders utilize the simple categorization of terms as either positive or negative. However, working with black-box approaches or inferring the overall valence of a disclosure merely from term frequencies is prone to error, since companies often frame negative news using positive words. 
	Our statistical procedure remedies this issue as it labels words based on their actual interpretation in financial materials. On the other hand, regulators can utilize our mechanism to put in place effective warning mechanisms for disclosures whose content can provoke critical market developments. 
		\item \textbf{Marketing.} Practitioners in the field of marketing strive to understand how people perceive language in advertisements and press releases. 
 Here, a granular understanding on a word-by-word basis would enable them to carefully consider phrasing in order to enhance sales. In addition, marketing teams could utilize our inference technique to make early predictions regarding the success of ad campaigns, product launches or the popularity of product attributes. In this vein, our method can identify words that influence customers in a positive or negative direction.
\end{itemize}\end{quote}
These examples highlight several prominent applications that benefit from a granular understanding of language at word level.

Ultimately, it is hoped that the contributions and advantages presented in this paper -- such as quantifying the reception of language -- will become an important tool in future research papers. Application of this method can yield novel insights into behavioral research questions regarding the information processing of natural language. This should help those in the field of social sciences to add to the growing body of knowledge on the role of behavior in individual decisions and population-wide outcomes, such as voting, consumer demand, information sharing, product evaluation and opinion aggregation. As demonstrated in this paper, our methodology has the potential to enable unprecedented opportunities in terms of validating behavioral research outside of existing laboratory setups. Yet it also fuels innovations in the theoretical advancement and formalization of theories as its high interpretive power facilitates new discoveries.

\subsection*{Summary}

Understanding the decision-making of individuals, enterprises and organizations presents a fundamental pillar of behavioral research. However, the challenges associated with processing natural language have been largely associated with simple decision models featuring predominantly structured data. Yet an unparalleled source of information is encoded in unstructured formats, and especially textual materials. The reasons behind this are multifaceted, including the recent advent of the big data era and the increasing availability of data through the World Wide Web, which has made a vast number of written documents -- such as user-generated content and news -- available to the public. 

Past research has laid the groundwork for inferring the polarity of written contents, albeit in a manner that is usually limited to a few psychological dictionaries that classify single terms. Such approaches work almost out-of-the-box and thus seem promising at first, but entail inevitable and major shortcomings. The elements of these word lists are selected \emph{ex ante} by manual inspection and subjective judgment. As such, our paper exposes the weaknesses of common dictionary methods: they only allow one to assess the overall polarity of documents and not of individual expressions, thereby leaving any deeper insights in the underlying text processing untapped. In addition, they often prove insufficient in adequately reflecting the domain-specific perception of a given audience.

As a remedy, this paper proposes the use of LASSO regularization as a form of variable selection to extract relevant words that statistically impact decisions. Social science researchers can greatly benefit from such a procedure, as it infers \emph{ex post} relevant terms based on the outcome of a decision. It can therefore efficiently adapt to domain-specific peculiarities of narratives and discriminate between subtle polarity levels across words. 

\section*{Acknowledgments}
The valuable contributions of Ryan Grabowski are gratefully acknowledged.


\bibliography{literature}

%

\end{document}